\def\year{2020}\relax
\documentclass[letterpaper]{article} % DO NOT CHANGE THIS
\usepackage{aaai20}  % DO NOT CHANGE THIS
\usepackage{times}  % DO NOT CHANGE THIS
\usepackage{helvet} % DO NOT CHANGE THIS
\usepackage{courier}  % DO NOT CHANGE THIS
\usepackage[hyphens]{url}  % DO NOT CHANGE THIS
\usepackage{graphicx} % DO NOT CHANGE THIS
\usepackage{graphicx}  % DO NOT CHANGE THIS
\title{Deep Poetry: A Chinese Classical Poetry Generation System}
\author{Yusen Liu,\thanks{Equal contribution.} Dayiheng Liu,\footnotemark[1] Jiancheng Lv\thanks{Correspondence to Jiancheng Lv. }\\
College of Computer Science, Sichuan University\\ %If you have multiple authors and multiple affiliations
24 Yihuan Road\\ Chengdu 610065, China\\
\{liuyusen96, losinuris\}@gmail.com, lvjiancheng@scu.edu.cn % email address must be in roman text type, not monospace or sans serif
}

\begin{document}

\maketitle

\begin{abstract}
In this work, we demonstrate a Chinese classical poetry generation system called Deep Poetry. Existing systems for Chinese classical poetry generation are mostly template-based and very few of them can accept multi-modal input. Unlike previous systems, Deep Poetry uses neural networks that are trained on over 200 thousand poems and 3 million ancient Chinese prose. Our system can accept plain text, images or artistic conceptions as inputs to generate Chinese classical poetry. More importantly, users are allowed to participate in the process of writing poetry by our system. For the user's convenience, we deploy the system at the WeChat applet platform, users can use the system on the mobile device whenever and wherever possible. The demo video of this paper is available at \url{https://youtu.be/jD1R_u9TA3M}.
\end{abstract}

\renewcommand{\thefootnote}{\arabic{footnote}}
\section{Introduction}
\noindent Chinese classical poetry plays a special and significant role in Chinese history. Poetry is the treasure of Chinese culture. During ancient times, many poets and poetry hobbyists were devoted to poetry and created a lot of poems. However, only a few best scholars can write coherent and beautiful poetry nowadays, most people either pay less attention to poetry or suffer from the high threshold of composing a poem. Therefore, we present a Chinese classical poetry generation system to arouse people's interest in poetry and make writing poetry easier.

Most recently, the automatic poetry generation has received great attention, neural networks have made this task a great development, including recurrent neural network \cite{zhang2014chinese}, encoder-decoder sequence to sequence model \cite{yi2017generating}, and neural attention-based model \cite{wang2016chinese}, etc. These studies are of great significance to entertainment and education.

However, most of the released Chinese poetry generation systems are template-based. The biggest drawback of this method is the generated poems are mostly monotonous because the templates restrict the content of the results. Moreover, these systems restrict users from entering text only and are unable to generate poetry according to images and artistic conceptions, such as the \textit{Daoxiangju} system\footnote{http://www.poeming.com} and the \textit{Microsoft Quatrain}.\footnote{http://duilian.msra.cn/jueju} Moreover, without the assistance of the teacher, it is difficult for users to compose poems using the systems above. To solve these problems, this paper introduces a new system that can accept plain text, images or artistic conceptions as input and allow users to participate in the process of generating. Our system not only can entertain people but also serve as a tool for poetry enthusiasts to make their works perfect. In other words, Our system plays the role of the assistant helping users polish their poems.

The \textit{jiuge} system\footnote{https://jiuge.thunlp.cn} is a human-machine collaborative poetry generation system using neural networks \cite{guo2019jiuge}. Comparing with the \textit{jiuge}, our model is different from them and our system uses a better way to help users write poems. Furthermore, rather than a web application, we deploy our system at the WeChat applet platform. WeChat is the largest social media app in China with over 1 billion monthly active users. This means that users can access our system easily and use it on the mobile device whenever and wherever possible.

In summary, the contributions of our Chinese classical poetry generation system are as follows: 1) Multi-modal input such as plain text, images or artistic conceptions can be accepted by our system. 2) Our system allows users to participate in the process of generating. 3) Our system is deployed at the WeChat applet platform for easy access.

\begin{figure*}
\centering
\includegraphics[width=0.95\textwidth]{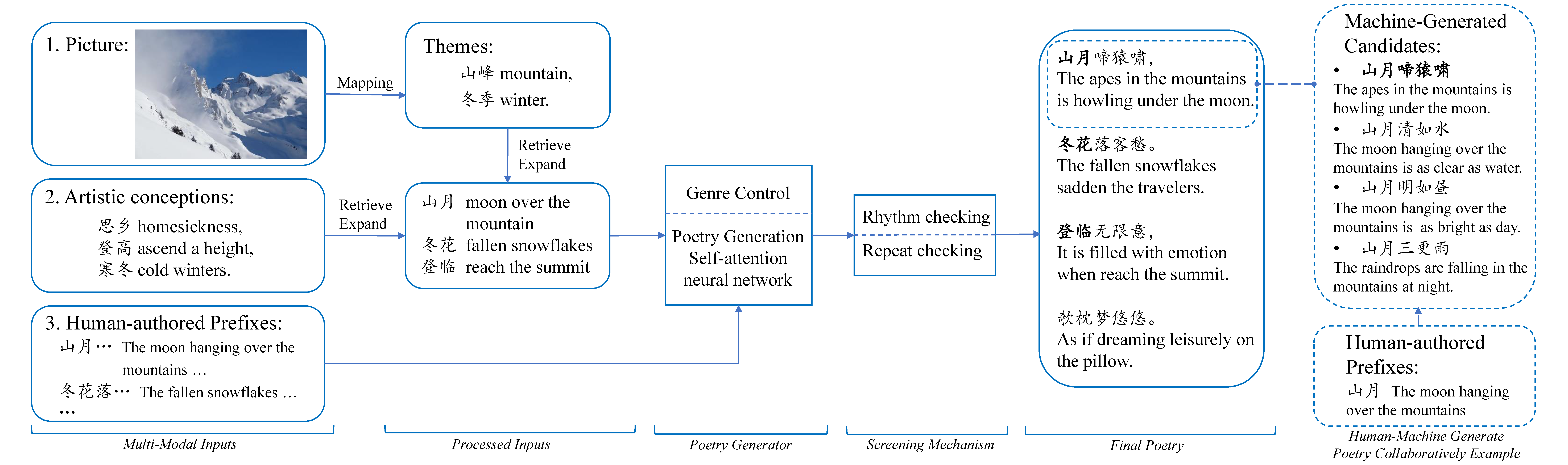}
\caption{The architecture of Deep Poetry system.}
\label{fig1}
\end{figure*}

\section{Architecture}
As shown in Fig. \ref{fig1}, the model which our system used consists of three components: a powerful method to process input, a self-attention neural network to generate poetry, and a screening mechanism for results. The model is mainly based on our previous work \cite{liu2018multi}.

Our system allows multi-modal input then generating poetry according to the input. For each image, we map it into at least two themes which recognized by Clarifai. The Clarifai API offers image recognition as a service. Then we retrieve and expand all related phrases. The phrases are from \textit{ShiXueHanYing}, which is a poetic phrase taxonomy. Each theme in \textit{ShiXueHanYing} contains dozens of related phrases. If the inputs are artistic conceptions, we will retrieve and expand all related phrases directly. Moreover, when users cooperate with our system to write poems, they can input sentences of any length and the system will generate some candidates using the human-authored prefixes as input.

The second component is a generator that can automatically generate poetry according to the processed input. A self-attention neural network is used as our poetry generator. The embedding of each phrase that retrieves from the previous step is fed into the neural network. Moreover, our system can control the genre, we provide several genres to users: quatrain with five or seven characters and acrostic poetry. For acrostic poetry, the generator will be attached to a condition that the first word of each line is fixed. Last but not least, the poetry will be generated word by word and a softmax logistic loss is employed to train the network.

The last component is a screening mechanism for results that generate with beam search (beam size=N). We design a mechanism to check these N candidates and remove the bad results. If the generated poems break the rhyming rules or have repetitive words, they will be classified as bad results by the screening mechanism and be abandoned.

\section{Demonstration}
The main functions of our system are generating poetry with multi-modal input, assisting users to write poetry and a word puzzle about poetry. For the first part, the system can generate poetry automatically using the plain text, images or artistic conceptions as input. Moreover, our system can generate acrostic poetry, which is a special genre in Chinese poetry. Users give somebody's name or short plain text to our system, then our system generates a poem using each word of the input as the first word of each line. After the poetry completes, it can be transferred to an exquisite card, the users can download it and share it with others.

As for the second part, our system allows users to write a satisfying poem with our system collaboratively. For each line, the users can write as long as they want, maybe a character, a word, or a sentence. The system will complete the whole line according to the prefix that the users have given. These generated candidates can serve as a reference for users and help them complete their poems. It is meant for people who are suffering from being unable to write a poem. After four lines, our system can generate a title according to the poem. For the word puzzle part, the system rearranges the words in popular poems and users should put them in the right order. In general, the game can improve user viscosity and arouse people's interest in classical poetry.

\section{Conclusion and Future work}
We propose Deep Poetry, a multi-function Chinese classical poetry generation system in this demo. Our system not only accepts multi-modal input but also allow users to participate in the process of writing poems, it is of great significance to both ordinary and professional users. Moreover, we add a word puzzle game to arouse public interest in traditional culture. without the hassle of installing the application program, our system can be easily accessed at the WeChat applet platform. In future work, we intend to integrate more genres generation into our system.

\section{Acknowledgments}
This work is supported by the National Key R\&D Program of China under contract No. 2017YFB1002201, the National Natural Science Fund for Distinguished Young Scholar (Grant No. 61625204), and partially supported by the Key Program of National Science Foundation of China (Grant No. 61836006).

\begin{small}
\bibliographystyle{aaai}
\bibliography{258-References}
\end{small}
\end{document}